# Universität Trier

# Morphological analyzer and generator for Pali

## Bachelor Thesis

for attainment of the academic degree of

Bachelor of Arts

at the University of Trier

Department of Digital Humanities and Computational Linguistics

presented by

**David Alfter**

**1032665**

**s2daalft@uni-trier.de**

Trier, January 2014

1st supervisor: Prof. Dr. Caroline Sporleder

2nd supervisor: Prof. Dr. Reinhard Köhler

# Abstract

This work describes a system that performs morphological analysis and generation of Pali words. The system works with regular inflectional paradigms and a lexical database. The generator is used to build a collection of inflected and derived words, which in turn is used by the analyzer. Generating and storing morphological forms along with the corresponding morphological information allows for efficient and simple look up by the analyzer. Indeed, by looking up a word and extracting the attached morphological information, the analyzer does not have to compute this information. As we must, however, assume the lexical database to be incomplete, the system can also work without the dictionary component, using a rule-based approach.

# Table of Contents

# 1. Introduction

## 1.1.    What is Pali ?

Pali (also written Pāli, Paḷi or Pāḷi) is a dead language from the group of Middle Indo-Aryan languages (Burrow, 1955: 2). Despite its status as dead language, Pali is still widely studied because many of the early Buddhist scriptures were written in Pali (Bloch, 1970: 8). It is also said that Buddha himself spoke Pali or a closely related dialect (Pali Text Society; Thera, 1953: 9).

Pali is an agglutinating and fusional language (Aikhenvald, 2007: 4). Besides a base meaning expressed by a stem or root, further information is expressed by adding affixes (Aikhenvald, 2007:4). However, in contrast to purely agglutinating languages, affixes in fusional languages cannot be split into separate meaning-bearing units; there are no clear-cut boundaries between the morphemes (Aikhenvald, 2007:4). Rather, a single affix expresses multiple morphemes, such as case, number and gender (Aikhenvald, 2007:4).

In the word *devo* 'a god' for example, the lemma is *deva* with the ending -o expressing NOUN, DECLENSION TYPE A, SINGULAR, MASCULINE, NOMINATIVE.

## 1.2.    Sanskrit

Pali is related to the more widely studied Indo-Aryan language *Sanskrit* (Duroiselle, 1921: 1). While Sanskrit was taught to members of the upper social class, Pali belonged to a group of vernacular dialects (Burrow, 1955: 36, 50). Buddha allegedly rejected Sanskrit as a language for his teachings because of its status as *learned language*, language of the upper classes, as opposed to vernacular dialects, spoken by the majority of people (Burrow, 1995: 58).

Much of the available literature on Pali grammar expect the reader to have at least some knowledge of Sanskrit, as Sanskrit is often used not only to trace the etymology of Pali words, but also to explain the grammar itself (Collins, 2006: ix). Duroiselle, in his *Practical Grammar of the Pali Language*, tries to abstain from using Sanskrit as an explanation for Pali, as he intended his grammar for students who do not know any Sanskrit (1921: 2). Even so, he concedes to sometimes using Sanskrit examples to explain seemingly idiosyncratic word formations in Pali (Duroiselle, 1921: 2).

Given the relationship between Pali and Sanskrit, quite some problems arising from the wish to computationally process Sanskrit are also valid for Pali. In the same vein, some linguistic methods and solutions for Sanskrit can be transferred to Pali. Some of the problems listed by Huet are the representation of phonemes, suitable transliteration schemes for program-intern processing and *sandhi*, sound changes occurring at morpheme boundaries (2005; 2009, as cited in Kulkarni and Shukl, 2009: 1-2).

The representation of phonemes did not pose any problems. However, due to the widespread use of Pali, it was also written in a myriad of scripts. We have, in this work, opted for a diacritical Unicode notation, based on the notation used by the dictionary of the Pali Text Society, on which the lexical database used in this work is based. Handling Sandhi is another important problem, and, while not exhaustive, has also been addressed in this work.

### 1.3. Motivation

Currently, there are no natural language processing (NLP) systems available for Pali. Thus, analyzing Pali texts is restricted to manual analysis. Even though there are some programs that allow the user to specify a word and generate  morphological forms (see, for example the *Pali Lookup System* by Aukana Trust[1]), there currently is, to the best of our knowledge, no system available that covers all the functionality presented in this work.

Having access to morphological information is important for all natural language processing systems (Kulkarni and Shuckl, 2009: 6). Many NLP tasks require the input to be at least Part-of-Speech-tagged. Indeed, many NLP systems are realized as pipelines, with the output from one module being the input for the next module. We are working on the start of this pipeline. Most systems for morphological analysis work on pre-tagged words (see, for example Lezius, Rapp and Wettler, 1998), rely on large lexical databases or a combination of both (see, for example Perera and Witte, 2005). Indeed, using morphological information in such systems can greatly increase accuracy (Silva, 2007: 42).

We can, however, not work with tagged words, since there are no reliable taggers available for Pali. Furthermore, the relation between morphological analyzers and taggers seems to be a circular one: morphological analyzers work best with tagged words, and taggers work best with morphologically analyzed words. Tagging and morphological analysis are two different NLP tasks but they support each other. One solution to this circular problem would be to use a tagged corpus and train a tagger on it. Unfortunately, there are virtually no tagged corpora available for Pali; manual tagging is a time-consuming and inefficient solution (Perera and Witte, 2005: 1).

In addition, the Pali-English dictionary by the Pali Text Society, the basis for the lexical database, presents some further problems. First of all, the dictionary does not present its information in an easily accessible way. This dictionary can be regarded as *the* standard, since there are only a handful of dictionaries for Pali. We would like to computationally process this corpus. The notation used in the dictionary is not always consistent, and even though the desired information is present, it is difficult to extract this information computationally. Another problem is that we cannot assume the dictionary to be complete. Lastly, we cannot assume the dictionary to be error-free. It is thus not recommended to rely solely on the dictionary.

---

[1] Available from http://metta.lk/pali-utils/

The aim of this work thus is to assist the user by providing tools for the morphological analysis of Pali. These modules should be useable by themselves or as a basis for higher-level systems such as Part-of-Speech taggers.

## 2. Challenges

Most grammars dealing with Pali refer to Sanskrit. These grammars expect the reader to have knowledge of Sanskrit and deduce Pali words from Sanskrit words and vice versa. Even without a presupposed knowledge of Sanskrit, performing morphological analysis and devising a system capable of automatically processing a language requires a certain level of understanding. As we had not been exposed to Pali prior to this work, we had to work ourselves into the language as quickly as possible.

Another challenge is the lack of literature on Pali. Working with as little, sometimes contradictory or incomplete, resources can be problematic. As Sanskrit presents similar problems as Pali with regard to morphology, using literature about Sanskrit seems to be a sensible choice. However, Sanskrit is written in Devanagari, an Indian script. Subsequently, literature dealing with the morphological analysis of Sanskrit does so using this script, which further complicates matters.

As a singularity, no new texts in Pali emerge. Therefore, the vocabulary of Pali does not change, as is the case with living languages. Given these circumstances, the occurrence of out-of-vocabulary words would be expected to be relatively low. The problem is that we have a non-computer-readable dictionary of yet unknown quality. We are in the process of processing this dictionary and making it computer-readable. However, the quality of the data would have to be checked.

Compounding is frequent and complex in Pali, following similar rules as Sanskrit. Furthermore, it is improbable that each and every single possible morphological form permissible in the language has at some point been written down. Therefore, assuming only the attested forms as correct is not a viable choice. We must allow all possible forms, including non-existent and possible erroneous forms when generating morphological forms.

The system here presented could theoretically work without a dictionary. The ultimate goal, however, is to continually improve the dictionary. With this goal in mind, dictionary functionality has been incorporated into the morphological analyzer and generator.

## 3. Lexical Database

The modules in this system access a lexical database. The implementation and specification of the database are not part of this work. The database holds different collections of words, also referred to as dictionaries. The collections of importance in this work are the *wordforms* collection, containing all occurring tokens from the corpus it was compiled from; the *lemma* collection, which contains only lemmata and, if available, word class information; and the *generated* collection, which contains data computed by this system's generator.

## 4. Generator

The core module of this work is the generator. The generator returns all possible regular inflected forms of a lemma. Section 4.1 explains the importance of the generator for this work. Section 4.2 gives a more detailed description of the approach taken. Finally, section 4.3 highlights some problems encountered.

### 4.1. Rationale

The approach taken in this work is as follows: given a set of lemmata, generate and store all inflected and derived forms of each lemma. Storing these generated forms in a separate dictionary inside the lexical database (*generated collection*), we can, at runtime, simply look up each encountered word during lemmatization. If we can find the encountered form inside the generated dictionary, the lemma can simply be retrieved by lookup. It must however be borne in mind that the set of lemmata (*lemma collection*) is in all probability not exhaustive. As such, the lemmatizer will have to rely on additional methods. This point will be elaborated on further in section 6 (Lemmatizer).

Furthermore, if, for any given word, the syntactic category (noun, verb,...) cannot be determined with certainty, the generator could be used. Indeed, generating all forms under the assumption that the current word is a noun or a verb and checking the occurrence of these forms against the database of actually occurring forms (*wordforms collection*) can be helpful. For example, if we have to choose between *verb* and *noun:* if many forms generated according to the nominal paradigm can be found, and forms generated according to the verbal paradigm cannot be found or occur with low frequency, it is more probable that the current word belongs to the nominal category. This technique could be used by other programs using this system.

## 4.2.    Approach

The generator works with *paradigms.* These paradigms are data models and were manually compiled from different works (mainly Duroiselle, 1921 and Collins, 2006) and then represented in XML format. A paradigm as data model is a hierarchical structure with the innermost node representing regular morphological endings, and all other traversed nodes to reach this node the morphological information encoded by this node. This data is later read and processed to yield paradigms as program *data structure*.

Let us consider the following excerpt from the pronominal paradigm model:

```
<pronoun>
    <personal>
        <number type="singular">
            <person type="1">
                <case type="nominative">
                    <ending>ahaṃ</ending>
                </case>
                <case type="accusative">
                    <ending>maṃ</ending>
                </case>
                ...
            </person>
        </number>
    </personal>
</pronoun>
```

We have two innermost nodes, namely `<ending>ahaṃ</ending>` and `<ending>maṃ</ending>`. To reach the first node (`ahaṃ`), we have to traverse the nodes *pronoun, personal, number, person, case*. This node thus expresses the morphological information: personal pronoun, first person singular nominative. The node `maṃ` expresses: personal pronoun, first person singular accusative.

Generally, the generator takes a lemma and a paradigm data structure and generates all forms according to the paradigm by first deriving the word-class-specific root of the word and then combining the root with the morphemes specified by the paradigm. In special cases, such as the pronominal paradigm, the generator determines the exact paradigm(s) the lemma belongs to (i.e. personal pronouns) and simply returns all forms from this paradigm.

## 4.3.    Problems

One problem of the generator is the lengthy and extensive computation it has to perform. If we want our system to be exhaustive, we have to allow for over-generation. If for example, a word does not have attached morphological information, the theoretical worst case scenario would consist of generating all forms according to the nominal, adjectival, verbal and numeral paradigm. In practice, the worst case scenario encompasses only the nominal, adjectival and numeral paradigms. If we assume the nominal paradigm to have two numbers, three genders, eight cases and only one ending per case, we arrive at 76 possibilities. If we assume the adjectival paradigm to have two numbers, three genders, eight cases and only one ending per case, we arrive at 76 possibilities. If we assume the numeral paradigm to have two numbers, no gender, eight cases and only one ending per case, we arrive at 16 possibilities. If we assume the prefix number to be at 24, with prefixes being attachable to all generated words, we arrive at a total of $(76 + 76 + 16) * 24 = 4032$ generated word forms from one single word. This case does occur with lemmata ending in *a* with no further information attached. In reality, however, there often is more than one case per ending and the total of generated words is in excess of 5000 for a word without morphological information.

Another problem is that the paradigms encode *regular* inflectional forms. Even though some irregular nouns and numerals have been added as separate temporary files, the generator will have to be extended with more functionality relating to irregular inflections.

## 5. Analyzer

The analyzer returns morphological information about incoming words, including the possible lemma(ta). There are two ways to analyze a word: dictionary- based and rule-based. Dictionary -based analysis consists in simply looking up the word in the dictionary of generated word forms. If a match is found, the corresponding entry is returned. If there is no corresponding entry in the dictionary, the analyzer falls back to the rule-based analysis. Section 5.1 will succinctly explain the dictionary-based lookup and section 5.2 will give an overview over the rule-based approach.

### 5.1. Dictionary-based approach

If we presuppose the dictionary collection of generated forms to be completed (see above), the dictionary will contain entries, such as (line breaks and indentation added for reading ease):

```
{
"word":"devāya",
"grammar":
        {
        "morphology":
            {
            "lemma":"deva",
            "information":
                {
                "paradigm":"noun",
```

```
            "gender":"masculine",
            "number":"singular",
            "case":"dative",
            "declension":"a"
            }
        }
    }
}
```

If we want to analyze the word *devāya,* we look up the word in the dictionary. In this example, we assume the word to be in the dictionary. We can then simply retrieve this entry, which contains the morphological analysis, as returned by the generator. If the dictionary lookup fails, the analyzer falls back to the rule based approach.

## 5.2. Rule-based approach

Given the grammatical regularity of Pali, we opted for a rule-based default approach. The analyzer takes a word and, if available, the word's word class. As in most cases, however, no Part-of-speech information can be supplied, a dedicated module responsible for guessing the word class(es) of a word has been created. Thus, if no information about the word class of a word can be supplied, the word's word class is guessed.

The analyzer then checks whether the word is a pronoun or an irregular word. Irregular word files have been included for offline functionality. However, this information should eventually be incorporated into the dictionary. If a match is found at this point, the morphological information is retrieved by look up and returned.

Otherwise, the analyzer tries to separate prefixes from the word and identifies the stem and ending of the word. Based on this putative stem and the word class information, whether given or guessed, the analyzer derives the lemma. The morphological information corresponds to the identified ending's morphological information. The full morphological analysis returns a lemma, morphological information and a separation of morphemes. The separation of morphemes is not available for pronouns and irregular words.

## 6. Lemmatizer

In contrast with Novák's terminology, the lemmatizer's function is to return the lemma and word class of a given word, excluding additional morphological information (Novák, 2004: 65). The functionality of what Novák calls *lemmatizer* is taken over by our morphological analyzer. However, this is but a terminological difference.

As we have seen in the previous section, the morphological analyzer already derives lemmata. This functionality is strongly dependent on morphological information, and cannot easily be isolated and moved from the analyzer to the lemmatizer. Therefore, calling the lemmatizer's `lemmatize` function in offline context results in the analyzer being invoked. The lemmatizer then extracts the lemma and word class of the result and returns this information.

In contrast, calling `lemmatize` in online context results in the word being looked up in the dictionary of generated word forms. As each of these entries contains the lemma from which it was derived, the lemma can simply be retrieved. If no matching entry can be found, the lemmatizer, as the analyzer, falls back to the offline method.

## 7. Sandhi Splitter

### 7.1. Sandhi

When words in Pali are combined, they often undergo phonological changes broadly known as *euphony* (Duroiselle, 1921: 6). More specifically, these sound changes are referred to as *sandhi*, the same term used as in Sanskrit (Duroiselle, 1921: 6). The word *sandhi* itself means *union* (Duroiselle, 1921: 6). Not unsurprisingly, Sanskrit sandhi rules can be mapped to Pali sandhi rules via another set of rules, and vice versa. However, the set of sandhi rules for Pali is smaller than the set of rules for Sanskrit.

There are different types of sandhi. Duroiselle divides them into *vowel sandhi, mixed sandhi* and *niggahīta sandhi* (1921: 6). To give a full overview over the sandhi rules would exceed the scope of this work, therefore I will give but a few examples to illustrate the different types of sandhi.

Vowel sandhi are sound changes that take place when two vowel sounds meet (Duroiselle, 1921: 6). One subtype of vowel sandhi is *elision.* In this case, either the first or the second vowel sound is elided (Duroiselle, 1921: 7).

ajja + upposatho = ajjuposatho (elision of first vowel sound)

cakkhu + indriyaṃ = cakkhundriyaṃ (elision of second vowel sound)

Mixed sandhi are sound changes that take place when a word ending in a vowel is followed by a word beginning with a consonant (Duroiselle, 1921: 12). One subtype is *consonantal reduplication*. In this case, the consonant is doubled.

pa + kamo = pa**kk**amo

vi + payutto = vi**pp**ayuto

Finally, niggahīta sandhi takes place when a word ending with a niggahīta (ṃ) is followed by a word starting with either a vowel or a consonant (Duroiselle, 1921: 14). The niggahīta may remain unchanged, is assimilated by the following letter, assimilates the following letter, is elided, leads to elision of the following letter or a combination of these processes.

taṃ + patto = taṃ patto (niggahīta remains unchanged)

saṃ + *m*ato = sammato (assimilation of niggahīta to *m*)

saṃ + *y*ogo = saññogo (assimilation of *y* to niggahīta and change from *ṃṃ* to *ññ*)

kiṃ + *i*ti = kinti (elision of *i* and assimilation of niggahīta to *t*)

As can be seen from the examples given, undoing these sound changes is not a trivial task. To illustrate some of the difficulties, let us look at elisions. The nature of the elided sound cannot be determined with certainty, or in other words, we cannot say which sound was elided. Furthermore, there is no straightforward way of identifying places where elision could have taken place. Therefore, undoing elisions amounts to generating all possible combinations for all potential elisions.

## 7.2.     Preliminary notes

The rules for the Sandhi Splitter described in this work were extracted from Duroiselle's *A Practical Grammar of the Pali Language (1921)*. In the first step, the rules, as listed by Duroiselle, were manually extracted. This step yielded rules like "A vowel followed by another vowel results in the elision of the first vowel". The next step consisted in rewriting these rules in a computer-interpretable way. I have opted for a *regular expression* syntax with back-references, constants and operations. The above mentioned rule thus becomes "(VOWEL) (VOWEL):$2". To better understand the structure of these rules, I will shortly digress and give a succinct explanation of the constructs used.

### 7.2.1.  Capturing groups and back-references

In regular expressions, a capturing group is a group of characters that is treated as a single unit and that is saved in memory for later retrieval, either directly or by back-reference. Capturing groups are typically written by enclosing an expression in parentheses. A back-reference is a reference to such a capturing group. Back-references are typically numbered, the number referring to the group, which almost always corresponds to the group's position in the regular expression. Given the following regular expression:

(A|The) (cat) lives (here)

There are three capturing groups: (A|The), (cat) and (here). "lives" is not a capturing group, since it is not enclosed in parentheses. This expression will recognize the sentences "A cat lives here" and "The cat lives here". When the regular expression is evaluated against either one of

these sentences (it will fail against other sentences), the following back-references will be created:

$1: A/The*

$2: cat

$3: here

* Back-reference $1 refers to either "A" or "The", depending on which one was encountered in the sentence the expression was evaluated against.

### 7.2.2. Constants

Based on the extracted sandhi rules, constants have been introduced for recurring sets of characters. Doing this greatly enhances readability and maintenance of the rules and also helps formulating the rules in a more succinct manner. These constants are defined in a separate file using a specific prefix notation. It should be noted that the prefix notation is unrelated to regular expressions and is merely a construct introduced by me to easily identify non-rules. During program execution, the constants will be replaced by the corresponding set of characters. As an example, *vowels* are often found in rules. Therefore, we have a constant named *VOWEL*, defined as:

=VOWEL:a,i,u,e,o,ā,ī,ū

The prefix *equals* (=) indicates that this is a constant definition. The first part of the definition contains the name of the constant (VOWEL), the colon separates the first part of the definition from the second part of the definition. The second part of the definition indicates the corresponding set of characters, each character separated from the next character by a comma. As an example of increased readability and succinctness, let us look at the definition of CONSONANT:

=CONSONANT:k,c,ṭ,t,p,kh,ch,ṭh,th,ph,g,j,ḍ,d,b,gh,jh,ḍh,dh,bh,y,r,l,v,h,s,ṅ,ñ,ṇ,n,m,ṃ

It should be clear that using constants not only improves readability and maintenance, but it also reduces the probability of errors in rules.

### 7.2.3. Operations

Some sandhi rules are of the form "If a long vowel follows a consonant, shorten the vowel" or "If a non-nasal sound follows a mute sound, elide the mute sound and reduplicate the non-nasal sound". An elegant way of rewriting these rules in a computer-interpretable way seems to be:

```
(CONSONANT) (LONG_VOWEL):$1+short($2)

(MUTE) (NON_NASAL):+duplicate($2)
```

That is by using operations. In these cases, the operations would be "short(x)" to shorten a (long vowel) sound and "duplicate(x)" to reduplicate a consonant. In the prefix notation, operations are introduced by *plus* (+). The operations are declared in the same file as the constants, but the exact implementation of the operations has to be declared inside the program, since reading partial program code from a file and integrating it into a program is overly difficult.

## 7.3. Approach

To come back to the example given at the beginning of this chapter, it should now be clear what

```
(VOWEL) (VOWEL):$2
```

does. VOWEL is replaced by the corresponding set of characters when the program is run, and whichever characters were matched, the second group is taken as replacement for the two vowels. To illustrate this, let us consider some possible matches. The expression would match the following expressions:

e o

u i

ā ū

All of these expressions consist of (VOWEL) (VOWEL). The back-reference $2 would be *o, i* and *ū* respectively.

However, the rules, as given by Duroiselle, are applicable only in one direction, namely the direction of *merging*, that is two or more sounds are merged. On the other hand, *sandhi splitting* is required to perform the exact opposite operation. Therefore, a system to reverse the rules had to be devised.

### 7.3.1. Reversing rules

Manually reversing the rules is not a viable option in most cases. Not all rules present the same degree of difficulty with regard to reversal.

Atomic rules are easiest to reverse. Atomic rules are rules that do not contain back-references, operations or *expandables*. Expandables are expressions stating alternatives. These expressions can be expanded to yield a number of separate rules, each with one of the given alternatives. This point will be elaborated on in the next paragraph. The following rule is an atomic rule:

```
e a:aya
```

Reversing this rule is accomplished by simply reversing the left-hand-side and the right-hand-side of the rule, yielding:

```
aya:e a
```

Rules containing expandables are more tricky to reverse. Let us consider the following rule:

```
(a|ā) (i|ī):e
```

Simply reversing this rule would yield:

```
e:(a|ā) (i|ī)
```

This would correspond to the rule "If you encounter the letter 'e', replace it by **either** a **or** ā followed by **either** i **or** ī". The right-hand-side of this rule thus expresses alternatives. The rule cannot be directly applied as is and has to be expanded into a separate set of rules by resolving the expandables. The astute reader will have noticed that capturing groups on the left hand side of a rule become expandables on the right hand side of the rule. For each expandable, we will create *n* separate rules, each with one of the *n* alternatives stated by the expandable. In a first step, this would yield two rules:

```
e:a (i|ī)
e:ā (i|ī)
```

As the newly created rules still contain expandables, we will resolve the remaining expandables. This yields:

```
e:a i
e:a ī
e:ā i
e:ā ī
```

This new set of rules contains only atomic rules. We thus expanded one rule into a set of four rules.

Rules containing back-references need special attention as well. Let us consider the rule:

```
(VOWEL) (iha|agga):$1t$2
```

Simply reversing it would yield:

```
    $1t$2:(VOWEL) (iha|agga)
```

The problem here is that the back-references cannot reference to anything, since there are no capturing groups preceding the back-references. They are thus meaningless. To solve this problem, we have to identify all back-references and the groups the back-references refer to. We then have to swap these *reference–back-reference pairs* before reversing the rule. The rule then becomes:

```
    (VOWEL)t(iha|agga):$1 $2
```

Finally, rules containing operations need special treatment. Let us consider the rule:

```
    (DENTAL) (CONSONANT):+duplicate($2)
```

Reversing this rule after having swapped the references–back-references yields:

```
    +duplicate((CONSONANT)):(DENTAL) $2
```

This rule is problematic in more than one way. First of all, the back-reference $2 cannot reference anything because there is at most one capturing group: (CONSONANT). This group being an argument to the operation, it can be debated whether this can be considered a capturing group or not. However, this is not of importance here. Secondly, the rule states "If you encounter **the operation 'duplicate'** with a CONSONANT…". However, the operation itself will never be encountered; the result of the operation will be encountered. To solve this problem, we again create a set of separate rules from this rule: First, the argument of the function, if it is a constant, is replaced by the corresponding set of characters. Then, for each character *c* in the set of characters, the function is called with the character *c* and the (invalid) back-reference is replaced by the character *c*. The complete output of this step is too voluminous to list here. It should be sufficient to give a short illustrative example of this step's output. For the character *t*, we invoke the function *duplicate(t)* which yields *tt*, and we replace the back reference $2 by *t*. We then take the next character *p* and proceed as described, and so on.

```
    tt:(DENTAL) t
    pp:(DENTAL) p
```

The rule still contains the expandable (DENTAL), which will further bloat the result list. The complete result of this rule's reversal can be found in the appendix (13.1 Rule reversal).

All these steps have been automated and the results have been manually checked and corrected to ensure the quality and correctness of the rules. Some rules that rely on information not accessible at this point were excluded. Such rules are of the form "i or u followed by a noun in the vocative case". Part-of-speech information is still virtually non-existent at this point and would require preprocessing by a POS-Tagger. It is possible that such information will become available at a later stage.

### 7.3.2. Splitting compounds

The module responsible for splitting compound words into their constituents operates on the basis of the above mentioned reversed rules. The module takes a word and first identifies which rules can be applied at which positions inside the word. It does so by comparing the rules against the word and position at hand, iterating through the word from left to right. The information from this step is saved in a table.

To illustrate this procedure, let us consider a hypothetical word consisting of the letters *abcdef*. Let the hypothetical rules RULE 1 and RULE 2 be as follows:

```
b:x y      (RULE 1)

d:w v      (RULE 2)
```

The algorithm starts at position 0, which corresponds first letter of the word. Since no rule of the form

```
a:x y
```

can be found, the algorithm proceeds to the next position. At position 1, we encounter the letter *b*. RULE 1 states that *b* should be replaced by *x y*. Thus a table entry is created with the current position and the relevant rule. Since there are no further rules applicable at position 1, the algorithm proceeds to the next position. The next table entry is created for position 3. All further positions do not result in table entries, since no applicable rules can be found. The table thus contains these entries:

| Position | Rule |
|----------|--------|
| **1** | RULE 1 |
| **3** | RULE 2 |

The algorithm then steps through each entry of the table and generates a result by applying the rule specified by the entry at the specified position in the word. Each result self-validates itself. Self-validation results in the result becoming invalid if it contains words made up of improbable letter combinations. When querying the confidence of a split, the result calculates its confidence of the split by looking up each resulting word in the dictionary. Confidence $c$ is simply expressed as :

$$c = \frac{n}{N}$$

with N being the total number of words in the result after splitting and n being the number of words from N found in the dictionary.

At this moment, the Sandhi Splitter can only split a word once. However, it is desirable to extend this behavior to split a word more than once. We have therefore introduced a parameter *depth.* Depending on the parameter *depth*, the table is traversed again with the current result list. Indeed, *depth* specifies the maximal number of splits that should be executed, which correlates with the expected number of constituents in the compound. The introduction of this parameter seems indispensable, as in most cases the number of constituents of a compound is not known. Conceptually, a depth of *zero* would simply return the received word, as no splitting at all would have to be performed. At depth 1, the result of the above mentioned hypothetical rules applied to the hypothetical word would yield:

a**x y**cdef       (RESULT 1)

abc**w v**ef       (RESULT 2)

Indeed, at depth 1, each rule in the table is applied once, resulting in at most one split, or in other words two separate words. At depth 2, the result would be:

a**x y**cdef       (RESULT 1)

abc**w v**ef       (RESULT 2)

a**x y**c**w v**ef       (RESULT 3)

The results RESULT 1 and RESULT 2 are the same as with depth 1. However, the table entries are again applied to these results. RULE 1 however is not applied to RESULT 1, as RESULT 1 already resulted from the application of RULE 1. Similarly, RULE 2 is not applied to RESULT 2. After updating the positions to reflect the correct position inside the results, applying RULE 1 to RESULT 2 yields RESULT 3. Applying RULE 2 to RESULT 1 also yields RESULT 3. Since the result is the same in both cases, it is only retained once. In this example, any depth greater than 2 will yield the same result as depth 2.

### 7.3.3. Merging compounds

The module responsible for merging two or more words into one word according to the rules of Sandhi operates on the basis of the above mentioned rules. The module takes a list of words and performs pair-wise merging. Starting with the first pair, the module identifies applicable rules, if any, and applies these rules to yield one or more merged words. If no applicable rule can be found, the words are left unmerged. If more than two words are specified, the module merges the remaining words after the first pair with the results of the first pair one by one.

## 8. Stemmer

A simple stemmer has been included as well. The stemmer simply removes all endings from a word, returning the stem of a word. The stem is not identical to the lemma of a word. Stemming is much faster than lemmatizing; in most cases, the result is not a grammatically valid word though. Still, the reduction of words to a common stem might prove useful for some future applications.

## 9. Technical documentation

The system consists of different parts. The following diagram gives a broad overview of the main modules.

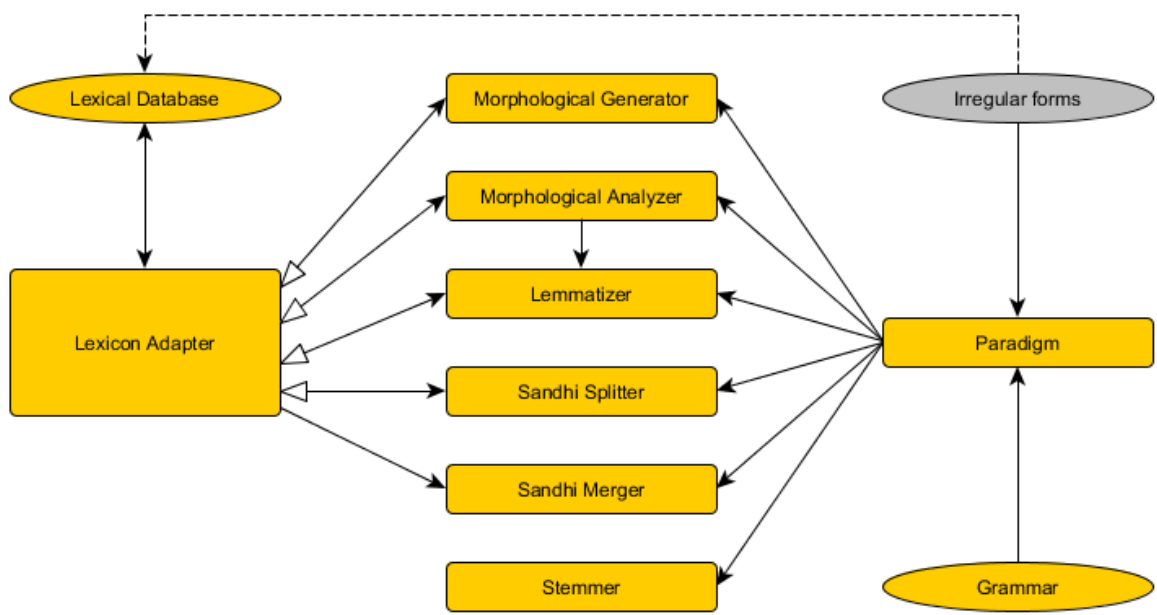

Ellipses stand for data. Rectangles stand for modules. Solid arrow heads denote an information flow from-to. Hollow arrow heads denote an optional information flow. Gray fields denote temporary, soon to-be-deprecated fields. A broken line stands for an anticipated permanent relocation of information from-to.

The implementation and specification of the lexical database are not part of this work. The lexical database is used to store different collections of words. The database is realized as a JSON-based database. A dedicated data management system has been created to facilitate the access to the database (to appear).

The diagram on the following page gives a more detailed picture of the interaction between the modules. The lexical database section has been left out to improve readability. Ellipses stand for data. Rectangles stand for modules. Hexagons stand for a collection of classes. For example, Strategy stands all Strategy classes (AdjectiveStrategy, NounStrategy, NumeralStrategy, NullStrategy,…).

Arrows in this diagram represent a *uses*-relation.

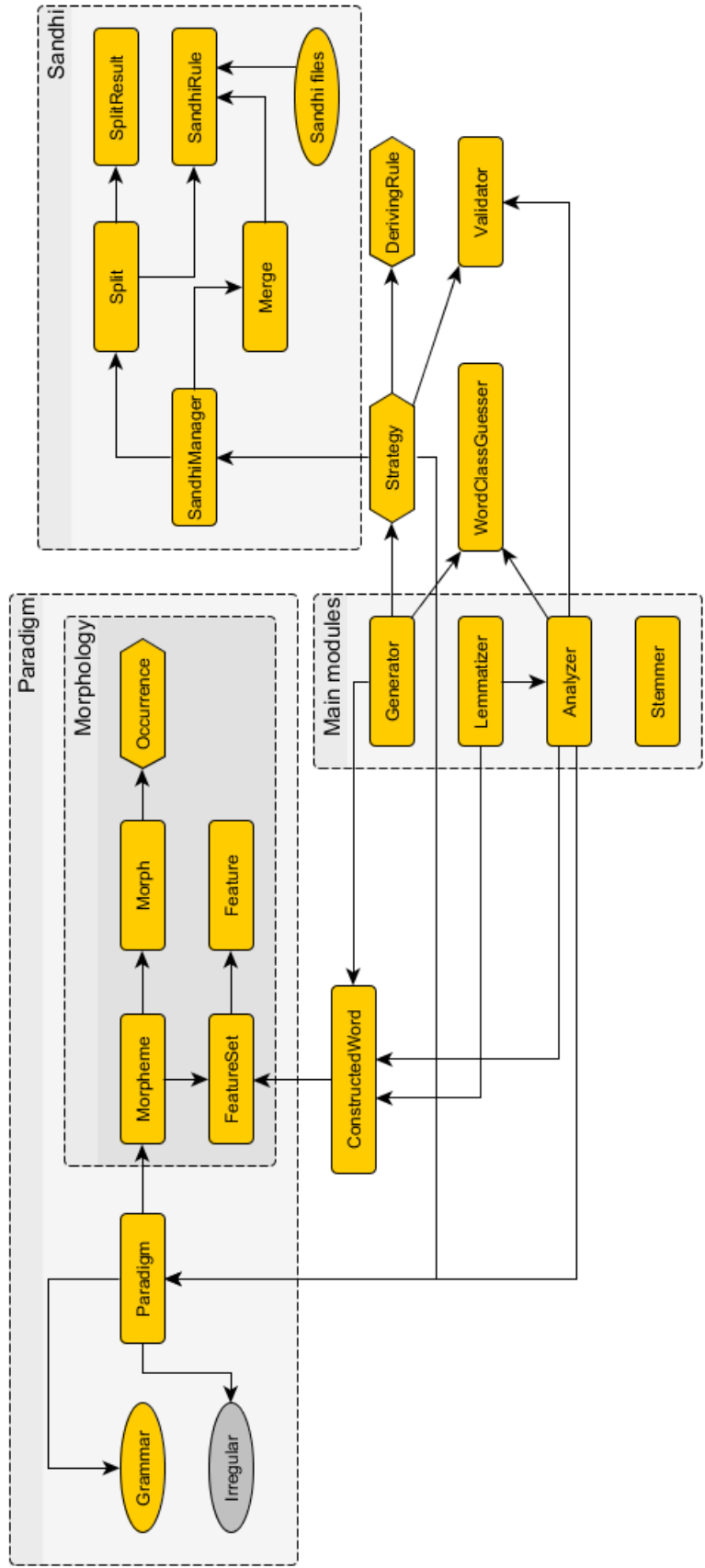

## 9.4. Data format and conventions

### 9.4.1. Input

#### 9.4.1.1. Grammar files

The grammar files are specified in XML format. The structure is hierarchical, with terminal nodes (innermost nodes) standing for particular morphs (word endings) or concrete word forms. The order of non-terminal nodes is not important. However, it must be ensured that all nodes traversed in order to reach a terminal node express that node's morphological information. In the following excerpt, we have two innermost nodes, namely <ending>aham</ending> and <ending>maṃ</ending>. To reach the first node (aham), we have to traverse the nodes *pronoun, personal, number, person, case*. This node thus expresses the morphological information: personal pronoun, first person singular nominative. The node maṃ expresses: personal pronoun, first person singular accusative.

```
<pronoun>
     <personal>
          <number type="singular">
             <person type="1">
                  <case type="nominative">
                       <ending>aham</ending>
                  </case>
                  <case type="accusative">
                       <ending>maṃ</ending>
                  </case>
                  ...
             </person>
          </number>
     </personal>
</pronoun>
```

If a node has an attribute, it is specified via the generic attribute name *type*.

#### 9.4.1.2. Temporary irregular files

The temporary irregular declension files are simple text files, with each line representing one entry. The following line is an example of an irregular declension file line.

eko{paradigm=numeral, number=singular, gender=masculine, case=nominative}

The first word is the morphological form, followed by the information encoded as key-value pairs. The key-value pairs are enclosed in curly braces. The key from a key-value pair is separated from the corresponding value by an equals sign. Key-value pairs are comma-separated.

### *9.4.1.3. Sandhi files*

Sandhi files are separated into rule files and dictionary files. Rule files contain one rule per line. Rules consist of a left hand side and a right hand side, separated by a colon. Both left and right hand side of a rule may contain literals and operations. Additionally, the left hand side may contain groups of literals, constants and regular expression elements. The right hand side may contain back-references.

Literals are characters that are to be matched as-is. Groups of literals are literals separated by the pipe character and enclosed in parentheses:

(j|c)

Operations start with a plus sign, followed by the function name, followed by the operation argument enclosed in parentheses. The operations have to be declared in the dictionary file, but have to be implemented as functions in the program itself. In the dictionary file, the definition of an operation is the plus sign, the operation's name, an opening parenthesis, the argument, a closing parenthesis:

+long(x)

Constants are named sets of characters. The constant name is written in all-caps. The equivalent set of characters has to be declared in the dictionary file. In the dictionary file, the definition of a constant is an equals sign, followed by the constant's name, followed by a colon, followed by the set of characters:

=VOWEL:a,i,u,e,o,ā,ī,ū

When used in rules, constants are not prefixed with an equals sign. When used in rules, constants must always be enclosed in parentheses.

Regular expression elements include the *start of line* ^ and the *end of line* $.

Back-references can be used to refer to a prior match by number. Back-references can only be used on the right hand side of a rule to refer to a grouping expression on the left hand side of the rule. Grouping expressions are enclosed in parentheses. The first grouping expression is the first expression enclosed in parentheses when reading from left to right. Back-references are written as the dollar sign followed by the number of the expression. Numbering of the expressions starts with 1. The number cannot be greater than the number of grouping expressions.

The dictionary file(s) contain the constant definitions and operation definitions.

All files may include end-of-line comments. Comments start with the hash character and extend to the end of the line. In-line comments are not supported; each comment must begin at the beginning of a line.

This convention was chosen because such rules, after some transformations, can be used by Java's *replaceAll* method, which takes a regular expression as first parameter, specifying the part to replace, and an expression as second parameter, specifying the expression to replace the first part with. The second parameter may contain back-references as declared by this convention.

### 9.4.2. Output

The lemmatizer, analyzer and generator return a List of JSON objects; a JSON object is an object with properties. Properties are expressed as key-value pairs. Property values can be queried by property name/key. All returned objects contain a key:

- "word", the value being the morphological form of a word.

- "grammar.morphology.lemma", the value being the lemma of "word"

- "grammar.morphology.information", the value being a list of key-value pairs that express the word's morphological information

JSON object was chosen as output format because it can directly be inserted into the lexical database. Furthermore, JSON object as a key-value map presents itself as an easily accessible model.

Internally, the program represents words that have been constructed by the morphological generator in objects of type *ConstructedWord*. Such constructed words are collection and then later converted to a suitable format for inserting into the dictionary. The latter is done by using the class *DictWord* provided by the library that provides a simple API to access the dictionary. ConstructedWord contains a feature set (key-value mappings) and the fields *lemma*, *word* and *stem*. The structure of a ConstructedWord is flat, compared to the nested DictWord structure.

The sandhi merger returns a List of String. The stemmer returns a String.

The sandhi splitter returns a List of SplitResult. SplitResult is a List of String containing the result of splitting a word, a list of SandhiTableEntries containing information about which rule has been applied at which position, and the confidence of the split result. SplitResult could have been converted to DictWord. However, the calculation of the confidence is done in a lazy manner. Calculating confidence is a lengthy computation and will only be performed when the confidence is required. Converting all SplitResults to DictWords would require the computation of every result's confidence.

### 9.5.     Paradigm

A Paradigm saves the information read from the grammar file for program-intern processing. The information from the grammar file is parsed into a morphological structure. Terminal nodes from the grammar file are saved as Morph objects, containing Occurrence information if present. If there is more than one terminal node at one level, the nodes are saved as a list of Morph objects inside a Morpheme object. Otherwise, the Morpheme object only contains a single Morph. The Morpheme object also saves the morphological information, that is all non-terminal nodes traversed so far, in the form of a FeatureSet containing a list of Feature objects (key-value pairs). Paradigm contains a list of Morpheme objects.

Occurrence objects represent information about the Morph it is attached to; Occurrence objects may contain information about how a Morph objects influences the context it occurs in.

### 9.6.     Generator

The generator takes a lemma as String, the lemma's word class as String and an optional list of options as input. Options are used for instance to specify the lemma's gender in the case of nouns.

If word class is not specified, the generator uses a word class guesser to guess the possible word classes. For lemmata, the word class depends solely on the ending of the lemma.

The generator uses a strategy manager to get the word class specific declension strategies and applies them to the lemma.

#### 9.6.1. Word class guesser

The word class guesser takes a word as String as input.

The word class guesser is responsible for returning possible word classes given a word of unknown word class. There are two different implementations of the guessing algorithm:

- If the word in question is known to be a lemma, the word class solely depends on the current ending of the word. The algorithms checks known lemma endings against the ending of the word at hand and returns all applicable word classes.
- If the word in question is not a lemma, the word class guesser checks all paradigms, trying to identify word class specific suffixes on the word. The algorithms counts the number of possible identified suffixes per word class paradigm. The algorithms weighs the counts, so that longer identified suffixes get more weight than shorter suffixes. The result is a weighted list of word classes. The word class with the most weight is, in most cases, the correct word class. Based on a pruning parameter, the list of possible word classes is cut off if the difference between the current weighted word class frequency and the weighted frequency of the following word class is less than the pruning parameter. A pruning parameter of 10 has been found to perform reasonably well.

### 9.6.2. Strategies

Strategies are responsible for deriving the word class specific stem of a word and combining the stem with the word class specific endings. A strategy retrieves the relevant paradigms and applies all relevant paradigms to the stem.

The GeneralDeclensionStrategy is a general strategy. It takes a lemma, a paradigm and a rule describing how to derive the stem from the lemma.

The AdjectiveStrategy, NounStrategy, NumeralStrategy and VerbStrategy are pre-configured classes that select the correct parameters for a given input and call the GeneralDeclensionStrategy with these parameters. The AffixStrategy is responsible for returning all possible combinations of prefixes and suffixes with a given word.

The GeneralDeclensionStrategy checks the validity of the generated forms using a validator. The validator uses simple, general rules to determine whether a word is valid or not. If a word is deemed invalid, the declension strategy tries to merge the stem and ending resulting in the invalid word using sandhi rules.

## 9.7.    Analyzer

The analyzer takes a word as String or DictWord and an optional list of options as input. Options are used to specify the word class of the input.

In offline context, the analyzer guesses the word class if it was not provided. The analyzer then checks whether the word is any form of a pronoun. If this is the case, the analyzer constructs a new analysis from the pronoun form and the attached morphological information and adds this analysis to the output list. The analyzer then checks whether the word is irregular. If this is the case, the analyzer constructs a new analysis from the morphological information attached to the irregular form and adds this analysis to the output list. The analyzer then tries to identify prefixes. For each guessed word class, the analyzer then tries to identify suffixes and paradigm endings. The analyzer then determines the boundary between the stem and the attached ending. If the ending contains declension information, this information is used in conjunction with the word class and putative identified stem to derive the word class specific lemma. The analyzer then constructs a new analysis using all the gathered information and adds this analysis to the output list. The analyzer then returns the output list.

The analyzer contains two overloaded methods for analyzing: *analyze* for offline context and *analyzeWithDictionary* for online context. These methods are overloaded to work with String and DictWord input.

In online context, the analyzer checks whether the input word is in the dictionary of generated word forms. If this is the case, the analyzer retrieves and returns all relevant entries. Otherwise, the analyzer falls back to offline mode.

## 9.8. Lemmatizer

The lemmatizer takes a word as String or DictWord as input.

In offline context, the lemmatizer calls the analyzer and extracts the relevant information.

In online context, the lemmatizer checks whether the input word is in the dictionary of lemma forms. If this is the case, the lemmatizer retrieves and returns this word wrapped as a singleton list of DictWord. Otherwise, the lemmatizer falls back to offline mode.

The lemmatizer contains two overloaded methods for lemmatizing: *lemmatize* for offline context and *lemmatizeWithDictionary* for online context. These methods are overloaded to work with String and DictWord input.

## 9.9. Sandhi splitter

The sandhi splitter takes a word as String and a depth as integer as input.

The sandhi splitter takes a word and first identifies which rules can be applied at which positions inside the word. It does so by comparing the rules against the word and position at hand, iterating through the word from left to right. The information from this step is saved in a table.

The algorithm then steps through each entry of the table and generates a result by applying the rule specified by the entry at the specified position in the word. Each result self-validates itself. Self-validation results in the result becoming invalid if it contains words made up of improbable letter combinations. Each result has a confidence. Confidence is only calculated when required. Confidence is expressed as the number of words of the split found in the dictionary divided by the total number of words in the split.

Depending on the parameter *depth*, the table is traversed again with the current result list. Indeed, *depth* specifies the maximal number of splits that should be executed, which correlates with the expected number of constituents in the compound. The introduction of this parameter seems indispensable, as in most cases the number of constituents of a compound is not known.

## 9.10. Sandhi merger

The sandhi merger takes two or more words as String array as input.

The sandhi merger merges two or more words. Internally, the words are merged pair-wise. The merger takes two words and identifies rules that are applicable to the word boundaries created by the ending of word 1 and the start of word 2 and merges these two words. The results of this merge operation are merged with the remaining words in the same manner, using each word from the result list as word 1 and the next of the remaining words as word 2, creating a new result list. This process continues until no words remain.

## 9.11.   Stemmer

The stemmer takes a word as String as input.

A simple stemmer has been included as well. The stemmer simply removes all endings from a word, returning the stem of a word. The stem is not identical to the lemma of a word. Stemming is much faster than lemmatizing; in most cases, the result is not a grammatically valid word though. The stemmer retrieves all paradigm endings, discarding all other attached information. The stemmer recursively strips off endings from a word until the word contains no strippable endings anymore. The remaining stem is returned.

## 9.12.   API

A high-level application programming interface (API) has been implemented to provide easy access to the main functions of the system. The API contains static methods that call the relevant system modules. To use the Pali NLP system via the API, simply use PaliNLP from the package de.unitrier.daalft.pali.

For example, to lemmatize, call:

```
PaliNLP.lemmatize("gavassa");
```

Alternatively, you can call the modules directly. The main modules are `Lemmatizer`, `MorphologyAnalyzer`, `MorphologyGenerator`, `NaiveStemmer` and `SandhiManager`. For example, to generate morphological forms using the MorphologyGenerator, call:

```
MorphologyGenerator.generate("go");
```

For a complete example of a simple program using the PaliNLP system, see Appendix 13.2.

# 10.   Future work

In the future, it would be desirable to further increase the functionality and accuracy of the presented system. Some areas could be improved by the inclusion of metathesis, dropping of syllables and epenthesis. Though some regular cases of metathesis and epenthesis are covered by the modules, not every case of metathesis or epenthesis could be covered. Another are that could need improvement is the Sandhi splitter. Compound words can only be split once by the Sandhi splitter at the moment. Even though this might be sufficient to resolve the majority of compounds, a more exhaustive splitting module will probably be necessary.

Another desirable evolution would be the development of a POS-Tagger for Pali, using the presented system as a basis. As mentioned earlier, POS-Tagging benefits from morphological information, and morphological analyzers benefit from POS-Taggers. This interdependency could be used to incrementally improve both systems as well as the quality and completeness of the dictionary.

Lastly, using n-grams to check the validity of words could be implemented and evaluated. The current system uses simple, general rules to determine whether a word is valid or not. However, using n-grams could possibly yield more accurate results and graded results, since anagram analysis would allow to check words at a finer granularity and assign a probability to the validity, correlating with the frequency of the constituting n-grams.

## 11. Conclusion

The presented system is a first step in the direction of the morphological analysis of Pali. The system is already functional, proving the concept to be viable and promising. As stated in the section 9, there is still some fine-tuning than can and should be done to improve this system, especially in the area of irregular declensions. Nonetheless, the system at the current stage of development should be able to process the majority of Pali words.

The system also uses a rather unconventional paradigm-based rule system to derive morphological information. Most morphological analyzers are built by using tools to derive a finite state transducer from a set of grammatical rules. This approach could have been applied in this case; it would have presupposed a better knowledge of the language though. However, given the regularity and immutability of Pali words, the rule-based approach seems reasonable.

Additionally, the incorporation of a lexical database allows for fast look up instead of a lengthy computation. The ultimate goal would be a system relying solely on the database. However, this would require the database to be reliable. Improving the quality of the lexical database would not only be reliant on the presented system, but also on a POS tagging system developed at a later stage. Still, this system can contribute to improving the existing lexical database.

As we have seen, Sandhi compounds still pose a problem. The presented system has made a first step towards resolving Sandhi-merged compounds, yielding a module that merges words according to the rules of Sandhi as well. However, more work needs to be done in order to be able to accurately split compounds.

# 12.   Sources

# 13. Appendix

## 13.1. Rule reversal

Reversing and resolving `(DENTAL) (CONSONANT):+duplicate($2)`

```
kk:t k        cch:dh ch      ḍḍ:s ḍ        bbh:d bh       ss:l s
kk:th k       cch:n ch       dd:t d        bbh:dh bh      ss:s s
kk:d k        cch:l ch       dd:th d       bbh:n bh       ṅṅ:t ṅ
kk:dh k       cch:s ch       dd:d d        bbh:l bh       ṅṅ:th ṅ
kk:n k        ṭṭh:t ṭh       dd:dh d       bbh:s bh       ṅṅ:d ṅ
kk:l k        ṭṭh:th ṭh      dd:n d        yy:t y         ṅṅ:dh ṅ
kk:s k        ṭṭh:d ṭh       dd:l d        yy:th y        ṅṅ:n ṅ
cc:t c        ṭṭh:dh ṭh      dd:s d        yy:d y         ṅṅ:l ṅ
cc:th c       ṭṭh:n ṭh       bb:t b        yy:dh y        ṅṅ:s ṅ
cc:d c        ṭṭh:l ṭh       bb:th b       yy:n y         ññ:t ñ
cc:dh c       ṭṭh:s ṭh       bb:d b        yy:l y         ññ:th ñ
cc:n c        tth:t th       bb:dh b       yy:s y         ññ:d ñ
cc:l c        tth:th th      bb:n b        rr:t r         ññ:dh ñ
cc:s c        tth:d th       bb:l b        rr:th r        ññ:n ñ
ṭṭ:t ṭ        tth:dh th      bb:s b        rr:d r         ññ:l ñ
ṭṭ:th ṭ       tth:n th       ggh:t gh      rr:dh r        ññ:s ñ
ṭṭ:d ṭ        tth:l th       ggh:th gh     rr:n r         ṇṇ:t ṇ
ṭṭ:dh ṭ       tth:s th       ggh:d gh      rr:l r         ṇṇ:th ṇ
ṭṭ:n ṭ        pph:t ph       ggh:dh gh     rr:s r         ṇṇ:d ṇ
ṭṭ:l ṭ        pph:th ph      ggh:n gh      ḷḷ:t ḷ         ṇṇ:dh ṇ
ṭṭ:s ṭ        pph:d ph       ggh:l gh      ḷḷ:th ḷ        ṇṇ:n ṇ
tt:t t        pph:dh ph      ggh:s gh      ḷḷ:d ḷ         ṇṇ:l ṇ
tt:th t       pph:n ph       jjh:t jh      ḷḷ:dh ḷ        ṇṇ:s ṇ
tt:d t        pph:l ph       jjh:th jh     ḷḷ:n ḷ         nn:t n
tt:dh t       pph:s ph       jjh:d jh      ḷḷ:l ḷ         nn:th n
tt:n t        gg:t g         jjh:dh jh     ḷḷ:s ḷ         nn:d n
tt:l t        gg:th g        jjh:n jh      vv:t v         nn:dh n
tt:s t        gg:d g         jjh:l jh      vv:th v        nn:n n
pp:t p        gg:dh g        jjh:s jh      vv:d v         nn:l n
pp:th p       gg:n g         ḍḍh:t ḍh      vv:dh v        nn:s n
pp:d p        gg:l g         ḍḍh:th ḍh     vv:n v         mm:t m
pp:dh p       gg:s g         ḍḍh:d ḍh      vv:l v         mm:th m
pp:n p        jj:t j         ḍḍh:dh ḍh     vv:s v         mm:d m
pp:l p        jj:th j        ḍḍh:n ḍh      hh:t h         mm:dh m
pp:s p        jj:d j         ḍḍh:l ḍh      hh:th h        mm:n m
kkh:t kh      jj:dh j        ḍḍh:s ḍh      hh:d h         mm:l m
kkh:th kh     jj:n j         ddh:t dh      hh:dh h        mm:s m
kkh:d kh      jj:l j         ddh:th dh     hh:n h         ṃṃ:t ṃ
kkh:dh kh     jj:s j         ddh:d dh      hh:l h         ṃṃ:th ṃ
kkh:n kh      ḍḍ:t ḍ         ddh:dh dh     hh:s h         ṃṃ:d ṃ
kkh:l kh      ḍḍ:th ḍ        ddh:n dh      ss:t s         ṃṃ:dh ṃ
kkh:s kh      ḍḍ:d ḍ         ddh:l dh      ss:th s        ṃṃ:n ṃ
cch:t ch      ḍḍ:dh ḍ        ddh:s dh      ss:d s         ṃṃ:l ṃ
cch:th ch     ḍḍ:n ḍ         bbh:t bh      ss:dh s        ṃṃ:s ṃ
cch:d ch      ḍḍ:l ḍ         bbh:th bh     ss:n s
```

## 13.2.    Simple program

```java
import java.util.List;

import lu.cl.dictclient.DictWord;
import de.unitrier.daalft.pali.PaliNLP;
import de.unitrier.daalft.pali.phonology.element.SplitResult;

public class Demo {

    public void run (String word) {
        String stem = PaliNLP.stem(word);
        List<DictWord> lemmata = PaliNLP.lemmatize(word);
        List<DictWord> analyses = PaliNLP.analyze(word);
        System.out.println("The stem of " + word + " is " + stem);
        for (DictWord lemma : lemmata)
            System.out.println("A possible lemma of " + word + " is " +
                lemma.toString());
        for (DictWord analysis : analyses)
            System.out.println("A possible analysis of " + word + " is " +
                analysis.toString());
    }

    public void generate (String lemma) {
        // expects word class as second parameter
        List<DictWord> forms = PaliNLP.generate(lemma, null);
        for (DictWord form : forms) {
            System.out.println("A possible form of " + lemma + " is " +
                form.toString());
        }
    }

    public void split (String word) {
        // expects splitting depth as second parameter
        // only depth 1 supported at the moment
        List<SplitResult> split = PaliNLP.split(word, 1);
        for (SplitResult result : split) {
            System.out.println("A possible split of " + word + " is " +
                result.toString());
        }
    }

    public void merge (String... words)  {
        List<String> mergedWords = PaliNLP.merge(words);
        for (String mergedWord : mergedWords) {
            System.out.println("A possible merge is " + mergedWord);
        }
    }

    public static void main (String[] args) {
        Demo demo = new Demo();
        demo.run("gavassa");
        demo.generate("go");
        demo.split("sakideva");
        demo.merge("saki", "eva");
    }
}
```

### 13.3.    User manual

## 1. Getting started

**Important**: Please make sure that you have Java 7 installed before using this program. The program will not work with a prior version of Java.

You should have received a CD-ROM containing the Pali NLP system. Copy the contents of this CD-ROM to your computer (for example to C:\PaliNLP). You can skip the next step (2. Compiling source code).

Alternatively, the source code can be cloned from https://github.com/daalft/PaliNLP/. For help on how to clone a repository from github, please see the github manual.

## 2. Compiling source code

Before compiling, make sure that you have the Java Software Development Kit (Java SDK) 7 installed. Before compiling, make sure that you have Apache Ant installed. For help on how to install Apache Ant, please see the Apache Ant manual.

Open a new shell window/command-line window and navigate to the path containing the source code (the path containing the src and data folders). Run the command

```
ant
```
After a successful build, run the command

```
ant jar
```
If the operation succeeds, you will find a jar file and four platform dependent scripts (PaliConsole.bat, PaliGUI.bat, PaliConsole.sh, PaliGUI.sh).

## 3. Console mode

To start the console mode, double-click the PaliConsole script that corresponds to your system. On Unix systems, launch PaliConsole.sh; on Windows systems, launch PaliConsole.bat. If neither of these works on your system, set the classpath to include the created jar and the packages

2013-01-19_LibDictionaryClientRecompiled.jar
Jackson-annotations-2.2.3.jar
Jackson-core-2.2.3.jar
Jackson-databind-2.2.3.jar

from the folder data/extlib/. Launch the program by calling de.unitrier.daalft.pali.PaliConsole.

For example, if we are on the path containing the created jar file PaliNLP-20XXYYZZ, on a Windows system (spaces are indicated for clarity):

```
java[SPACE]-cp[SPACE].\PaliNLP-
20XXYYZZ.jar;.\data\extlib\jackson-databind-
2.2.3.jar;.\data\extlib\jackson-core-
2.2.3.jar;.\data\extlib\jackson-annotations-
2.2.3.jar;.\data\extlib\2014-01-
19_LibDictionaryClientRecompiled.jar[SPACE]de.unitrier.daalft.pal
i.PaliConsole
```

with 20XXYYZZ corresponding to the timestamp of the jar (for example PaliNLP-20140126.jar).

The PaliNLP console should open. The console mode has been specifically written to provide access to the PaliNLP system via the console.

Each command in the console should be followed by `[ENTER]`. For example, the instruction

Type

```
lemma
```

means that you should enter the word 'lemma', then press the `[ENTER]` key.

## 4. GUI mode

To start the graphical user interface (GUI) mode, double-click the PaliGUI script that corresponds to your system. On Unix systems, launch PaliGUI.sh; on Windows systems, launch PaliGUI.bat. If neither of these works on your system, set the classpath to include the created jar and the packages

2013-01-19_LibDictionaryClientRecompiled.jar
Jackson-annotations-2.2.3.jar
Jackson-core-2.2.3.jar
Jackson-databind-2.2.3.jar

from the folder data/extlib/. Launch the program by calling de.unitrier.daalft.pali.PaliGUI. For example, if we are on the path containing the created jar file PaliNLP-20XXYYZZ, on a Windows system (spaces are indicated for clarity):

```
java[SPACE]-cp[SPACE].\PaliNLP-
20XXYYZZ.jar;.\data\extlib\jackson-databind-
2.2.3.jar;.\data\extlib\jackson-core-
2.2.3.jar;.\data\extlib\jackson-annotations-
2.2.3.jar;.\data\extlib\2014-01-
19_LibDictionaryClientRecompiled.jar[SPACE]de.unitrier.daalft.pal
i.PaliGUI
```

with 20XXYYZZ corresponding to the timestamp of the jar (for example PaliNLP-20140126.jar).

The PaliNLP GUI window should open. The GUI mode has been specifically written to provide access to the PaliNLP system via a graphical user interface.

## 5. Lemmatizer

*In the console:*

If you have activated any mode other than the Analyzer, type

```
chmod
```

to return to the mode selection.

Type

```
lemma
```

Enter a word, for example:

```
buddhassa
```

You will get a list of possible lemmata. You can also specify the word class of a word by appending it to the word, using a colon as separator:

```
buddhassa:noun
```

*In the GUI:*

Enter one or more words in the input field. Multiple words should be separated by spaces.

Click the Lemmatize button. You will get a list of possible lemmata. You can also specify the word class of the word(s) by appending it to the word(s), using a colon as separator.

## 6. Stemmer

*In the console:*

If you have activated any mode other than the Stemmer, type

```
chmod
```

to return to the mode selection.

Type

```
stem
```

Enter a word, for example:

```
gavena
```

You will get the word stem. You can also specify the word class of a word by appending it to the word, using a colon as separator. However, this information does not influence the result.

*In the GUI:*

Enter one or more words in the input field. Multiple words should be separated by spaces.

Click the Stem button. You will get the word stem. You can also specify the word class of the word(s) by appending it to the word(s), using a colon as separator. However, this information does not influence the result.

## 7. Analyzer

*In the console:*

If you have activated any mode other than the Analyzer, type

```
chmod
```

to return to the mode selection.

Type

```
ana
```

Enter a word, for example:

```
gavena
```

You will get a list of possible analyses.

*In the GUI:*

Enter one or more words in the input field. Multiple words should be separated by spaces.

Click the Analyze button. You will get possible analyses. You can also specify the word class of the word(s) by appending it to the word(s), using a colon as separator.

## 8. Generator

Please note that generating word forms may take some time to complete.

*In the console:*

If you have activated any mode other than the Generator, type

```
chmod
```

to return to the mode selection.

Type

```
gen
```

Enter a word, for example:

```
go
```

You will get a list of morphological forms.

*In the GUI:*

Enter one or more words in the input field. Multiple words should be separated by spaces.

Click the Generate button. You will get a list of morphological forms. You can also specify the word class of the word(s) by appending it to the word(s), using a colon as separator.

# 9. Sandhi

## 9.4.  Splitting

Please not that splitting can take some time to complete.

*In the console:*

If you have activated any mode other than the Sandhi Splitter, type

```
chmod
```

to return to the mode selection.

Type

```
ss
```

Enter a word, for example:

```
sakideva
```

You will get a list of possible splits.

*In the GUI:*

Enter one or more words in the input field. Multiple words should be separated by spaces.

Click the Split button. You will get a list of possible splits. You can also specify the word class of the word(s) by appending it to the word(s), using a colon as separator. However, this information does not influence the result.

## 9.5.  Merging

*In the console:*

If you have activated any mode other than the Sandhi Merger, type

```
chmod
```

to return to the mode selection.

Type

```
sm
```

Enter two or more words separated by spaces, for example:

```
saki eva
```

You will get a list of possible merges. Please not that it is not possible to specify word classes when using the Sandhi Merger.

*In the GUI:*

Enter two or more words in the input field. Multiple words should be separated by spaces.

Click the Merge button. You will get a list of possible merges. Please not that it is not possible to specify word classes when using the Sandhi Merger.

## ERKLÄRUNG ZUR BACHELORARBEIT

Hiermit erkläre ich, dass ich die Bachelorarbeit selbständig verfasst und keine anderen als die angegebenen Quellen und Hilfsmittel benutzt und die aus fremden Quellen direkt oder indirekt übernommenen Gedanken als solche kenntlich gemacht habe.

Die Arbeit habe ich bisher keinem anderen Prüfungsamt in gleicher oder vergleichbarer Form vorgelegt. Sie wurde bisher nicht veröffentlicht.

_______________________________________     _______________________________________
Datum                                        Unterschrift